\documentclass[letterpaper]{article} 
\usepackage{aaai25}  
\usepackage{times}  
\usepackage{helvet}  
\usepackage{courier}  
\usepackage[hyphens]{url}  
\usepackage{graphicx} 
\usepackage{natbib}  
\usepackage{caption} 
\usepackage{algorithm}
\usepackage{algorithmic}
\usepackage{amsmath,amsfonts}

\usepackage{colortbl}
\usepackage[dvipsnames]{xcolor}
\usepackage{graphicx}
\usepackage{booktabs}
\usepackage{multirow}
\usepackage{amsmath}

\usepackage{newfloat}
\usepackage{listings}
\DeclareCaptionStyle{ruled}{labelfont=normalfont,labelsep=colon,strut=off} 
\floatstyle{ruled}
\newfloat{listing}{tb}{lst}{}
\floatname{listing}{Listing}
\title{Bridging Knowledge Gap Between Image Inpainting and Large-Area Visible Watermark Removal}
\author{
    Yicheng Leng\textsuperscript{\rm 1, \rm 2}, Chaowei Fang\textsuperscript{\rm 1}\thanks{Corresponding author.}, Junye Chen\textsuperscript{\rm 3}, Yixiang Fang\textsuperscript{\rm 2},  Sheng Li\textsuperscript{\rm 4}, Guanbin Li\textsuperscript{\rm 3, \rm 5}
}
\affiliations{
    \textsuperscript{\rm 1} School of Artificial Intelligence, Xidian University, Xi’an, China\\
    \textsuperscript{\rm 2} School of Data Science, The Chinese University of Hong Kong, Shenzhen, China\\
    \textsuperscript{\rm 3} School of Computer Science and Engineering, Research Institute of Sun Yat-sen University in Shenzhen, Sun Yat-sen University, Guangzhou, China\\
    \textsuperscript{\rm 4} Afirstsoft, Shenzhen, China\\
    \textsuperscript{\rm 5} GuangDong Province Key Laboratory of Information Security Technology\\
}

\usepackage{bibentry}

\begin{document}

\maketitle
\begin{abstract}
Visible watermark removal which involves watermark cleaning and background content restoration is pivotal to evaluate the resilience of watermarks. Existing deep neural network (DNN)-based models still struggle with large-area watermarks and are overly dependent on the quality of watermark mask prediction. To overcome these challenges, we introduce a novel feature adapting framework that leverages the representation modeling capacity of a pre-trained image inpainting model. Our approach bridges the knowledge gap between image inpainting and watermark removal by fusing information of the residual background content beneath watermarks into the inpainting backbone model. We establish a dual-branch system to capture and embed features from the residual background content, which are merged into intermediate features of the inpainting backbone model via gated feature fusion modules. Moreover, for relieving the dependence on high-quality watermark masks, we introduce a new training paradigm by utilizing coarse watermark masks to guide the inference process. This contributes to a visible image removal model which is insensitive to the quality of watermark mask during testing. Extensive experiments on both a large-scale synthesized dataset and a real-world dataset demonstrate that our approach significantly outperforms existing state-of-the-art methods. The source code is available in the supplementary materials.
\end{abstract}

\section{Introduction}
\label{sec:intro}
\begin{figure*}[t]
\begin{center}
    \includegraphics[width=0.7\linewidth]{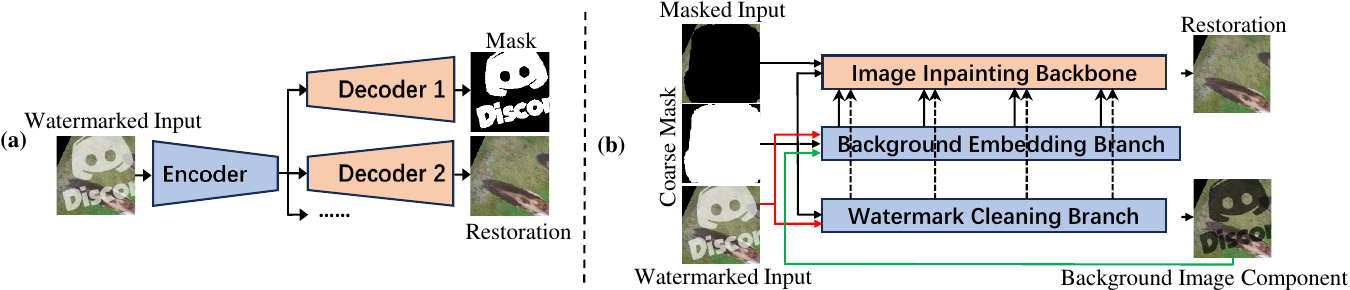}
\end{center}
   \caption{
   (a) Existing multi-task based methods such as~\cite{cun2021split,liang2021visible,sun2023denet} adopt a shared encoder and multi-branch decoder for implementing sub-tasks such as watermark segmentation, watermark decomposition, and background restoration; (b) We propose a novel solution through adapting an image inpainting backbone with prompt information extracted from a watermark component cleaning and a background content embedding branches. Moreover, we relieve the dependence on high-quality watermark masks by leveraging coarse masks to guide the inference process.}
   
   \label{fig1}
\end{figure*}



Visible watermarks serve a pivotal role in asserting image ownership and copyright. Yet, they can obscure vital content, especially during image editing or in situations where tampered information within images plays a crucial role. Studying on visible watermark removal contributes to evaluating the resilience of watermarks. This paper aims to develop a deep neural networks (DNN) based model which revolves around two objectives: watermark cleaning and background content restoration.

Prior visible watermark removal works, depicted in Figure~\ref{fig1} (a), embrace a multi-task framework that employs various decoder branches to implement sub-tasks such as watermark mask segmentation, watermark decomposition, and background content restoration.
Notable models like~\cite{hertz2019blind,cun2021split,liang2021visible,sun2023denet} are instrumental in this approach.
However, existing methods still face two challenges: 
1) In real-world images, areas of watermarks can be very large. Those large-area watermarks inevitably accentuate the complexity of background content recovery, especially when they are overlaid on regions with intricate visual content.
Due to insufficient representation modeling ability, the performance of existing models still have substantial improvement room when coping with large-area watermarks.
2) Since DNN-based models can usually overfit the training data, they are able to generate high-quality watermark masks on training images. This makes the learned models dependent to quality of the predicted watermark masks on real-world images. Missed detection leaves discernible watermark traces, while too many false alarms cause confusion to the background content recovery process.

In light of these challenges, we are dedicated in borrowing the rich knowledge of the image inpainting model to address the visible watermark removal task.
Recent advances in the image inpainting field demonstrate that DNN models are able to fill in missing regions of images with plausible visual content. Employing the knowledge of the image inpainting model to foster visible watermark removal model is a direction worth exploring.
However, distinct to the image inpainting task, the residual background content beneath watermarks can provide valuable prompts for background content restoration.
To bridge the knowledge gap between visible watermark removal and image inpainting, we propose a feature adapting framework which can effectively fuse the information of residual background content into intermediate features of a pre-trained image inpainting model, as depicted in Figure~\ref{fig1} (b). 
First, to capture the information of the residual background content residing in the watermarked input, we set up a watermark component cleaning branch which directly predicts an image precluding the watermark information from the input image. Moreover, we construct the other branch to further embed the background content.
The intermediate features of the above two branches can provide valuable prompt information for repairing the regions destroyed by watermarks.
Hence, we utilize gated fusion modules to merge features extracted by the two branches into intermediate features of the image inpainting backbone.
With help of the above feature adapting framework, we build up a novel watermark removal model which can combine the prior knowledge of image inpainting and prompt information of residual background content beneath transparent watermarks.

\begin{figure}[h]
\begin{center}
    \includegraphics[width=0.8\linewidth]{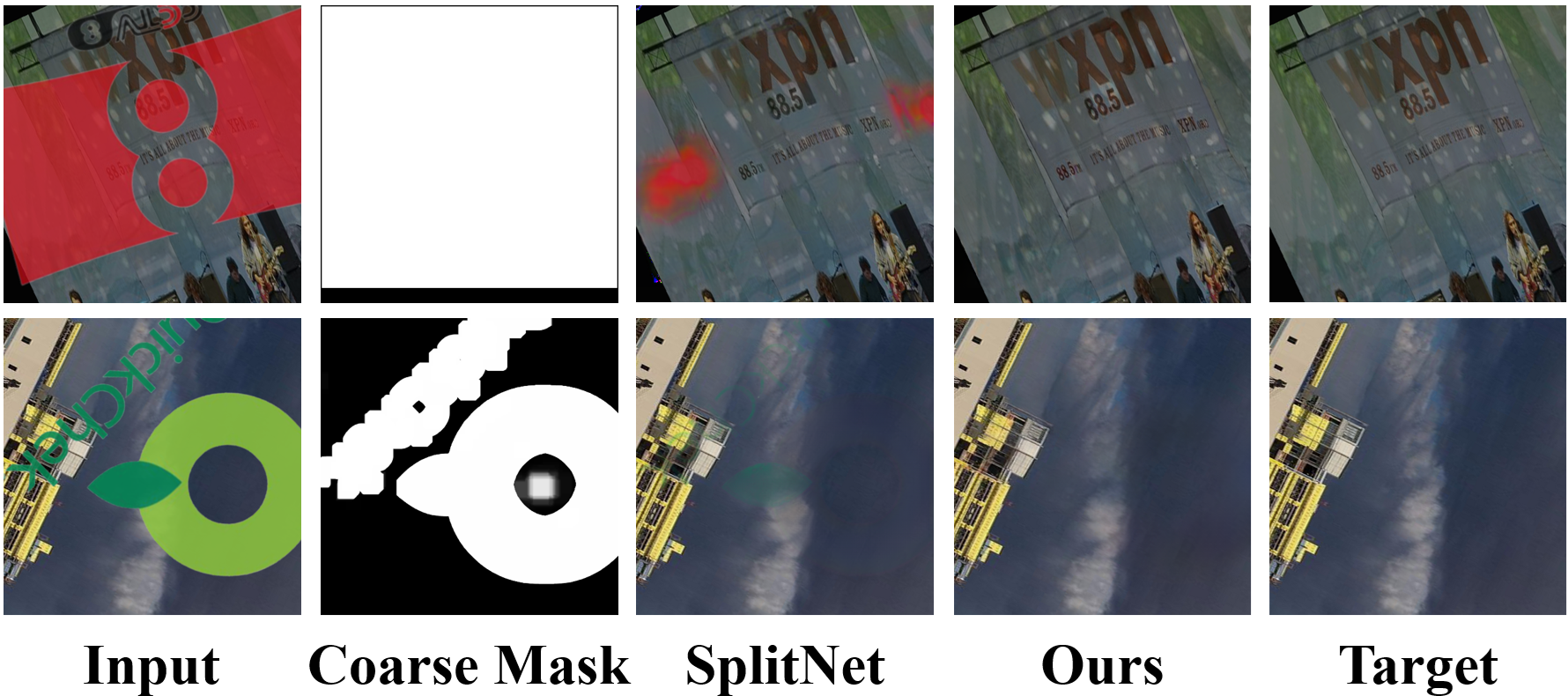}
\end{center}
   \caption{The first column showcases two input examples which are covered by large-area watermarks. Though coarse watermark masks (second column) are available, our method (fourth column) can effectively remove these watermarks and accurately recover the background, showing significant superiority over SplitNet (third column).
   }
\label{fig2}
\end{figure}

Awareness of the watermark region is critical to explore the knowledge of the image inpainting, while preventing the watermark component from affecting the background restoration process. However, the precise segmentation of watermark is also very challenging. In this study, we relieve the dependence on high-quality watermark detection performance by allowing coarse watermark masks to guide the image restoration process. Straightforwardly, we synthesize moderately corrupted watermark masks and integrate them with watermarked images as inputs. Such a training strategy helps foster a model insensitive to the quality of the segmented watermark masks. During practical usage, a rather coarse watermark mask is sufficient for the learned model to realize watermark removal. Figure~\ref{fig2} displays examples of watermark removal accomplished by our model.
We conduct extensive experiments on a large-scale synthesized dataset and a real-world dataset, and the results demonstrate that our proposed method achieves state-of-the-art performance.

Our main contributions are summarized as follows:
\begin{itemize}
\item We propose a novel feature adapting framework for tackling the large-area watermark removal problem, capable of combing the prior knowledge of a pre-trained image inpainting model and prompt information of residual background content. 
\item A new training paradigm is devised for improving the robustness of watermark removal model against low-quality watermark segmentation masks, enhancing the stability in coping with real-world watermarked images. 
\item Through comprehensive evaluations on a large-scale synthesized dataset and a real-world dataset, our method sets a new state-of-the-art benchmark for large-area visible watermark removal.
\end{itemize}

\section{Related Work}
\label{sec:related}
\subsection{Visible Watermark Removal}

Visible watermark removal involves restoring images that are covered by watermarks to their original watermark-free state. 
This task is challenging since watermarks have diverse shapes, areas, colors, and transparency levels.
Typical methods in this field rely on a multi-task pipeline. \citet{cheng2018large} learn an object detection model~\cite{redmon2017yolo9000} to locate watermarks and then construct a U-Net~\cite{ronneberger2015u} model for transforming watermarked input to watermark-free output.

\citet{hertz2019blind} propose a method that utilizes a shared encoder with separate decoders to predict the watermark image, watermark mask, and background image, enhancing watermark removal performance while maintaining low network complexity. Similarly, \citet{cun2021split} leverage task-specific attention mechanisms to create multiple decoder branches within a shared parameter space, reducing parameter redundancy. They also introduce a refinement stage to further improve restoration quality.
\citet{liang2021visible} adopt dual decoder branches for watermark mask prediction and background restoration, using the predicted watermark mask to guide feature extraction in the background restoration branch, effectively enhancing features in regions affected by the watermark. Additionally, \citet{sun2023denet} utilize contrastive learning with multi-head attention \cite{vaswani2017attention, dosovitskiy2020image} to disentangle watermark and background information.
Despite these advancements, existing methods struggle to restore images heavily corrupted by large-area watermarks and rely heavily on high-quality watermark masks. Given the strengths of image inpainting models in leveraging long-range context for image repair, our approach focuses on adapting pre-trained image inpainting models to tackle the challenges of large-area watermark removal.

\subsection{Image Inpainting}

Image inpainting techniques~\cite{bertalmio2000image} focus on filling missing parts of an image with content that matches the surrounding visual context, and many studies \cite{suvorov2022resolution,li2022mat,liu2018image,dong2022incremental,zuo2023generative,yang2023uni,liu2023coordfill} have explored this area extensively. These methods can be applied to visible watermark removal by treating watermark regions as missing parts. However, they fail to leverage the residual background beneath transparent watermarks. To address this limitation, we introduce a dual-branch design that adapts the inpainting model's intermediate features by cleaning watermark components and embedding background content.

\begin{figure*}[t]
\begin{center}
    \includegraphics[width=1\linewidth]{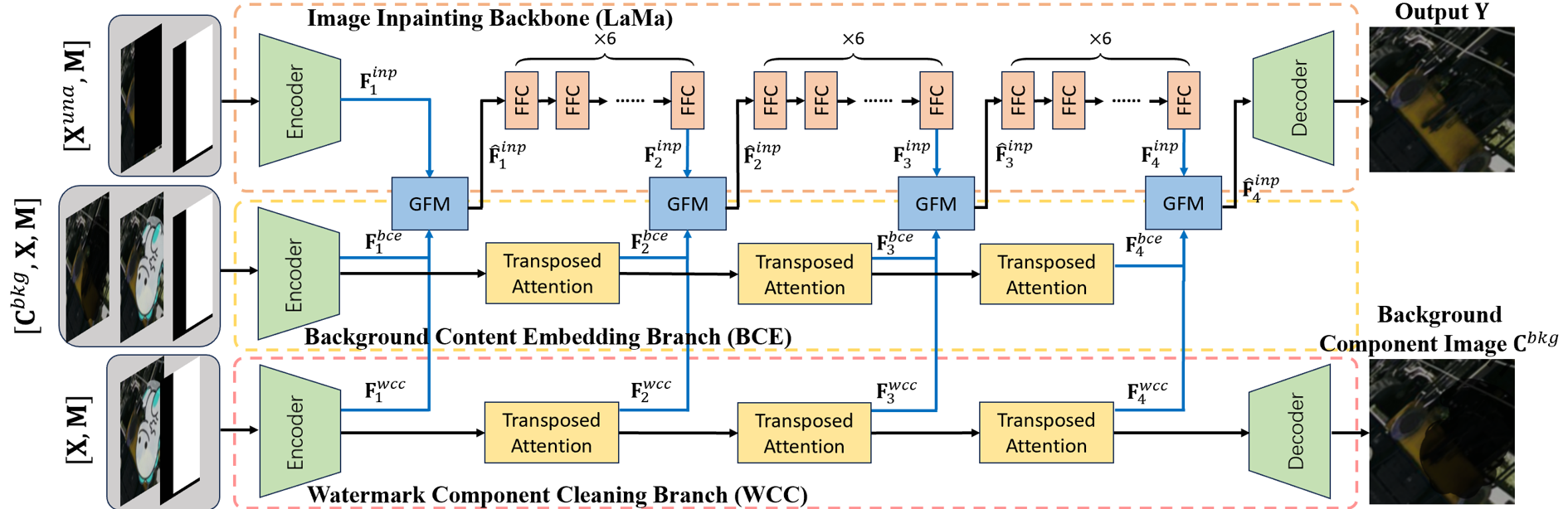}
\end{center}
   \caption{Overview of our framework which adapts an image inpainting backbone model, LaMa, to address the visible watermark removal task.  
   Given an input image $\mathbf X$ and a coarse mask $\mathbf M$, the watermark component cleaning branch (WCC) is employed to preclude the interference information brought by watermarks from the input image.
   Then, a background content embedding branch (BCE) is used to extract prompt features from the background component image and the original input image.
   We enhance the intermediate features of LaMa with these feature extracted by WCC and BCE branches.
   }
\label{fig3}
\end{figure*}

\section{Methodology}
This paper endeavors to develop a model with the capacity to convert a watermarked image into its watermark-free version, in which a coarse mask is given to specify the watermark region. 
We define the input watermarked image be $\mathbf X \in \mathbb R^{h\times w \times 3}$ and the coarse watermark mask be $\mathbf M \in \{0,1\}^{h\times w \times 1}$, where $h$ and $w$ represent the height and width of the input image, respectively. 

\subsection{Method Overview}

The key challenges in visible watermark removal are the thorough elimination of watermark components and the seamless reconstruction of the damaged background. Global context information is particularly critical for accurately restoring backgrounds obscured by large-area watermarks.
To address these challenges, we adopt the pre-trained image inpainting model LaMa~\cite{suvorov2022resolution} as our backbone due to its strong performance in reconstructing extensive masked regions and capturing global context through fast Fourier convolution (FFC)~\cite{chi2020fast}. This capability is especially advantageous for removing visible watermarks that obscure significant background details, ensuring the naturalness and authenticity of the restored content.
To further enhance LaMa’s performance, as shown in Fig.~\ref{fig3}, we introduce a watermark cleaning branch to remove watermark interferences from the input image and produce a cleaned background component image. Additionally, a background content embedding branch extracts features from both the original input and the cleaned background image to support the reconstruction of affected regions. The features from these branches are then fused with the LaMa backbone via gated fusion modules, ultimately generating the final restored background image.


\subsection{Watermark Cleaning and Background Embedding}
The image inpainting model effectively captures global context for reconstructing missing regions, but it overlooks crucial residual background information under the visible watermarks. To address this, we introduce a feature adapting framework that enhances the inpainting model's intermediate features using insights from watermark component cleaning and background content embedding branches.

\subsubsection{Watermark Component Cleaning (WCC) Branch}
Considering that the watermark component is irrelevant to the background content, we first establish a branch to subtract it from the input image. Since the information from the whole image is needed for recognizing watermarks, the long-distance relations become crucial. Therefore, we design the watermark component cleaning branch to effectively capture global context information.

As depicted in the bottom section of Figure~\ref{fig3}, regarding the concatenation of $\mathbf X$ and $\mathbf M$ as the input, an encoder with the same architecture of LaMa's encoder is utilized to extract a resolution-reduced feature map $\mathbf F^{wcc}_1 \in \mathbb R^{h^\prime \times w^\prime \times d}$.
Since the transposed attention module~\cite{zamir2022restormer} has outstanding advantages at extracting global context information, we stack three transposed attention modules to further enhance $\mathbf F^{wcc}_1$.
Suppose the outcome of the $i$-th transposed attention module be $\mathbf F^{wcc}_{i+1}$. 
The calculation process of $\mathbf F^{wcc}_{i+1}$ is summarized as follows:
\begin{itemize}
    \item [1)] A $1\times 1$ convolution layer and $3\times 3$ depth-wise convolution layer \cite{chollet2017xception} are employed to infer the query, key, and value variables, denoted by $\mathbf Q$, $\mathbf K$, and $\mathbf V \in \mathbb R^{h^\prime \times w^\prime \times d}$ respectively, from $\mathbf F^{wcc}_{i}$. The calculation process can be formulated as:
     \begin{small}
    \begin{equation}
    [\mathbf Q, \mathbf K, \mathbf V] = \mathtt{DConv_{3x3}}(\mathtt{Conv_{1x1}}(\mathbf F^{wcc}_{i})),
    \end{equation}
    \end{small}
    \item [2)] The horizontal and vertical dimensions of $\mathbf Q$ and $\mathbf K$ are unfolded into a single dimension, resulting in $\mathbf q$ and $\mathbf k \in \mathbb R^{(h^\prime w^\prime) \times d}$, respectively. Namely, $\mathbf q = \mathtt{unfold}(\mathbf Q)$, and $\mathbf k = \mathtt{unfold}(\mathbf K)$, where $\mathtt{unfold}(\cdot)$ represents the space-to-channel unfolding operation.
    Then, a cross-channel correlation map $\mathbf S$ is inferred by ($\alpha$ is a constant): 
    \begin{small}
    \begin{equation}
    \mathbf S = \mathtt{Softmax}(\mathbf q^{\mathtt{T}} \mathbf k / \alpha), 
    \end{equation}
    \end{small}
    \item [3)] Upon the calculation of $\mathbf S$, $\mathbf F^{wcc}_{i+1}$ is generated by: 
    \begin{small}
    \begin{equation}
    \mathbf F^{wcc}_{i+1}=\mathbf F^{wcc}_{i} + \mathtt{Conv_{1x1}}( \mathtt{fold} (\mathtt{unfold}(\mathbf V) \mathbf S) ), 
    \end{equation}
    \end{small}
    where $\mathtt{fold}(\cdot)$ denotes the channel-to-space folding.
\end{itemize}

At the end of this branch, a decoder expands the feature map resolution to generate a background component image, $\mathbf C^{bkg}$. This branch serves two key purposes: identifying residual background content and generating features essential for restoring the background image.

\subsubsection{Background Content Embedding (BCE) Branch }
To more effectively leverage the prompt information contained in the generated residual background content, we introduce a Background Content Embedding (BCE) branch. To address potential loss of background information by the background component cleaning branch, $\mathbf{X}$ and $\mathbf{M}$ are reused to enrich the input of $\mathbf{C}^{bkg}$.
As shown in the middle section of Figure~\ref{fig3}, the BCE branch comprises an encoder followed by three transposed attention modules. The resulting feature maps, denoted as $\mathbf{F}^{bce}_1$, $\mathbf{F}^{bce}_2$, $\mathbf{F}^{bce}_3$, and $\mathbf{F}^{bce}_4$, explicitly capture the information of the background content, which are paramount for background reconstruction.

\subsection{Backbone Model Adaptation}
\subsubsection{Introduction to Backbone Model}
The architecture of LaMa is composed of three stages: an encoder for extracting preliminary features; an intermediate feature enhancement module consisting of 18 FFC modules; and a decoder.

First, we remove the content inside the watermark mask $\mathbf M$ of $\mathbf X$, deriving $\mathbf X^{una} = (1-\mathbf M) \circ \mathbf X$, where $\circ$ denotes the broadcast Hadamard product.
Regarding the concatenation of $\mathbf X^{una}$ and $\mathbf M$ as the input, the encoder produces a feature map $\mathbf F^{inp}_1 \in \mathbb R^{h^\prime \times w^\prime \times d}$, where $h^\prime = h/32$ and $w^\prime = w/32$.

We divide the 18 FFC modules of the intermediate feature enhancement module into three groups with each containing six FFC modules. 
Define the outcome of the $i$-th group of FFC modules as $\mathbf F^{inp}_{i+1} \in \mathbb R^{h^\prime \times w^\prime \times d}$.
The information of the residual background content provides valuable hints for recovering the regions destroyed by watermarks.
To take advantage of such kind of information, we employ the features extracted by WCC and BCE branches to enhance each $\mathbf F^{inp}_i$ with help of gated fusion modules (GFM).
As shown by Figure~\ref{fig3}, every GFM module enhances one intermediate feature map with a pair features extracted by WCC and BCE. 
We denote the enhanced counterpart of $\mathbf F^{inp}_i$ be $\hat{\mathbf F}^{inp}_i$.
These enhanced feature maps are regarded as the input for next FFC module groups. 
The final enhanced feature map $\hat{\mathbf F}^{inp}_4$ is fed into the decoder, deriving the final output $\mathbf Y$.

\subsubsection{Gated Fusion Module}

The intermediate features from backbone lack modelling of the residual background information.
Therefore, we devise a gated fusion module (GFM) to combine features from WCC and BCE with output features from FFC module groups, inspired by \cite{jin2023dnf}.

\begin{figure}[t]
\begin{center}
    \includegraphics[width=0.7\linewidth]{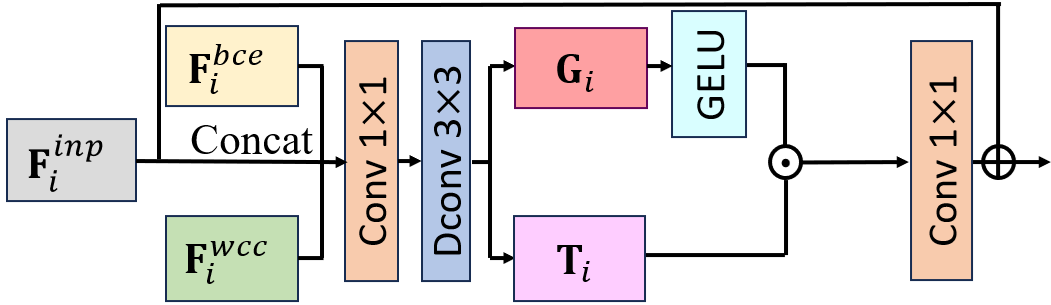}
\end{center}
   \caption{Illustration of the gated fusion module (GFM).}
\label{fig4}
\end{figure}

As shown in Figure~\ref{fig3}, four GFM-s are incorporated to enhance the intermediate feature maps $\{\mathbf F_i^{inp}\}_{i=1}^4$ of the image inpainting backbone.
The design of GFM is illustrated by Figure~\ref{fig4}.
The $i$-th GFM incorporates $\mathbf F_i^{wcc}$ and $\mathbf F_i^{bce}$ to adapt $\mathbf F_i^{inp}$.
First, a $1\times 1$ convolution layer and a $3\times 3$ depth-wise convolution layer are utilized to process the concatenation of $\mathbf F_i^{wcc}$, $\mathbf F_i^{bce}$, and $\mathbf F_i^{inp}$, resulting in a gate map $\mathbf G_i$ and a temporary feature map $\mathbf T_i$, namely,
\begin{small}
\begin{equation}
[\mathbf G_i, \mathbf T_i] = \mathtt{DConv_{3x3}}(\mathtt{Conv_{1x1}}([ \mathbf F^{wcc}_{i}, \mathbf F_i^{bce}, \mathbf F_i^{inp}] )).
\end{equation}
\end{small}
The final output of the $i$-th GFM is calculated through: 
\begin{small}
\begin{equation}
\hat{\mathbf F}_i^{inp} = \mathbf F_i^{inp}+\mathtt{Conv_{1x1}}( \mathtt{GELU}(\mathbf G_i) \circ \mathbf T_i),
\end{equation}
\end{small}
where $\mathtt{GELU}(\cdot)$ represents the Gaussian error linear unit function \cite{hendrycks2016gaussian}.

The GFM's effectiveness lies in its ability to highlight relevant features while suppressing less important ones. This is advantageous in our context, where different branches address complementary aspects of watermark removal. The GFM’s gating mechanism enables the model to concentrate on features crucial for reconstructing the watermarked region while filtering out irrelevant ones.

\begin{figure}[t]
\begin{center}
    \includegraphics[width=\linewidth]{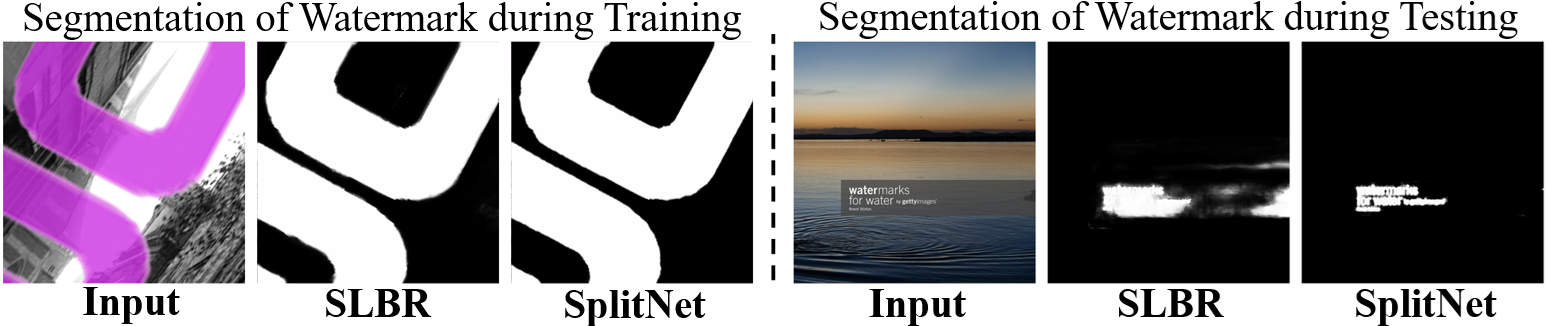}
\end{center}
   \caption{Watermark segmentation generated by blind visible watermark removal methods SLBR and SplitNet.
   }
\label{fig10}
\end{figure}

\subsection{Utilization of Coarse Watermark Mask}
Given the diverse appearances and shapes of watermarks, pinpointing their pixel-wise locations is challenging. Existing methods like SLBR and SplitNet, which use specific modules for watermark segmentation, struggle with generalization to real-world images, as shown in Figure~\ref{fig10}. To mitigate this issue, we use coarse watermark masks to guide the removal process. During training, we create these coarse masks by applying random erosion or dilation to ground-truth masks. For testing, we either manually outline the watermark region or use a segmentation model to generate the mask, which then guides the background restoration.

\subsection{Loss Function}
To balance various objectives, including pixel-level accuracy, perceptual quality, and reality, the designed loss function for training our model encompasses three types of constraints, inspired by \cite{suvorov2022resolution}. 
First, we use the $L_1$ loss to improve the pixel-level accuracy of predicted background component image $\mathbf C^{bkg}$ and the final restored watermark-free image $\mathbf Y$:
\begin{small}
\begin{equation}\label{eq9}
    L_{pixel} = || \mathbf Y - \mathbf G^{wf}||_1 + ||\mathbf C^{bkg}- \mathbf G^{bkg}||_1,
\end{equation}
\end{small}
where $\mathbf G^{wf}$ and $\mathbf G^{bkg}$ represent the ground-truth watermark-free image and background component image. 

The perceptual loss~\cite{johnson2016perceptual, zhang2018unreasonable} is also incorporated to improve the perceptual quality of $\mathbf C^{bkg}$ and $\mathbf Y$ based on semantic features extracted via the pre-trained ResNet50 model~\cite{he2016deep}.
The calculation formulation of this loss is as follows, 
\begin{small}
\begin{align}
    L_{per} =& \sum_{m=1}^M ( ||\mathtt{ResNet}^{(m)}(\mathbf Y)- \mathtt{ResNet}^{(m)}(\mathbf G^{wf}) ||_2 + \nonumber \\
    & ||\mathtt{ResNet}^{(m)}(\mathbf C^{bkg})- \mathtt{ResNet}^{(m)}(\mathbf G^{bkg}) ||_2 ),
\end{align}
\end{small}
where $M$ denotes the number of feature maps used for calculation, and $\mathtt{ResNet}^{(m)}(\cdot)$ produces the $m$-th feature map. 

Finally, the patch-wise adversarial training loss~\cite{isola2017image} is applied for improving the visual reality of $\mathbf Y$. Suppose the patch-wise discriminator model be $\mathcal D(\cdot)$.
The training loss for the discriminator is defined as follows:
\begin{small}
\begin{align}
    L_D = &-\Gamma(\log(\mathcal D(\mathbf G^{wf})))-\Gamma(\log ( \mathcal D(\mathbf Y) \circ (1-\mathbf M) ))- \nonumber\\
    & \Gamma(\log(1-\mathcal D(\mathbf Y))\circ \mathbf M),
\end{align}
\end{small}
where $\Gamma(\cdot)$ denotes the element summation operation.
The adversarial regularization for the watermark removal model is formulated as follows,
\begin{small}
\begin{align}
    L_G = -\Gamma(\log(\mathcal D(\mathbf Y))\circ \mathbf M ).
\end{align}
\end{small}

To avoid the gradient fluctuation brought by the adversarial training, we introduce the gradient penalty $P = ||\nabla_{\theta_G} L_G||^2_2$ where $\theta_G$ represents the parameters of the visible watermark removal model. Besides, an additional perceptual loss $L_{per}^\prime$ is calculated using features extracted by the discriminator. 

The total loss for training the watermark removal model is as follows:
\begin{equation}\label{eq13}
    L = \omega_1 L_{pixel}+\omega_2 L_{per}+\omega_3 L_G+\omega_4 L_{per}^\prime + \omega_5 P,
\end{equation}
where $\omega_1$, $\omega_2$, $\omega_3$, $\omega_4$, and $\omega_5$ are weighting factors for the above loss terms. 

\begin{figure}[b]
\begin{center}
    \includegraphics[width=\linewidth]{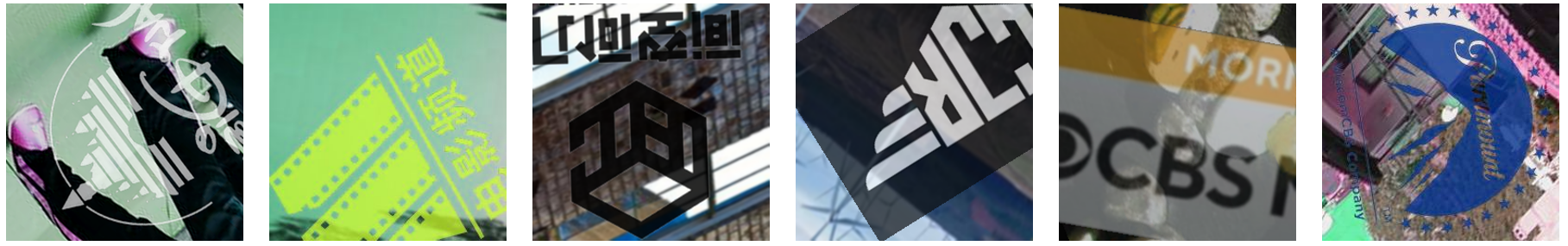}
\end{center}
   \caption{Watermarked images from our constructed dataset.
   }
\label{fig5}
\end{figure}
\section{Experiments}
\subsection{Datasets} 



\noindent\textbf{$\bullet$ ILAW.} 
Real-world images often feature large-area watermarks and undergo complex distortions such as compression or resampling, leading to the degradation of both watermark and background details. To advance research in large-area watermark removal, we introduce the \textit{Images with Large-Area Watermarks} (ILAW) dataset. The training set includes 60,000 images of size 256$\times$256 with 1,087 different watermarks, while the validation set contains 10,000 images of size 512$\times$512 with 160 distinct watermarks, different from those in the training set. Background images are sourced from the Places365 Challenge dataset \cite{zhou2017places}, and watermarks are collected from the Internet.
Given a clean background image $\mathbf I$ and a watermark image $\mathbf W$, we composite them into a watermarked image $\mathbf X$,
\begin{small}
\begin{equation}\label{eq1}
\mathbf X = \mathcal T((1 - \mathbf A) \circ \mathbf I + \mathbf A \circ \mathbf W),
\end{equation}
\end{small}
where $\mathbf A\in [0,1]^{h\times w\times 1}$ denotes alpha channel, and $\mathcal T(\cdot)$  represents random image distortion function consisting of image compression and resampling operations. 
The ground-truths of watermark-free background image and watermark-excluded background component image are obtained via $\mathbf G^{wf} = \mathcal T(\mathbf I)$ and $\mathbf G^{bkg} = \mathcal T((1 - \mathbf A) \circ \mathbf I)$.
For synthesizing the coarse watermark mask, we first generate a precise mask $\mathbf M_0$ from $\mathbf A$, i.e., $\mathbf M_0 = \mathbf A>0$.
Then, the same resampling operations used for generating $\mathbf X$ and a random dilation operation are adopted to process $\mathbf M_0$, resulting in $\mathbf M$. Examples of our dataset are displayed in Figure~\ref{fig5}, where the watermark area are much larger and more opaque than pictures in popular datasets used in watermark removal works.




\begin{table}[t]
  \centering
  \renewcommand{\arraystretch}{0.5}
  \setlength{\tabcolsep}{1mm}{ 
    \begin{tabular}{lccccc}
    \toprule
     Methods     & PSNR$\uparrow$  & SSIM$\uparrow$  & RMSE$\downarrow$  & RMSE$_w\downarrow$ & LPIPS$\downarrow$ \\
    \midrule
    \multicolumn{6}{l}{Performance with fixed mask per image} \\
    \midrule
    \rowcolor{gray!15}
    WDNet & 23.86  & 0.887  & 19.98  & 22.99  & 0.170  \\
    SplitNet & 25.90  & 0.901  & 15.99  & 19.15  & 0.147  \\
    \rowcolor{gray!15}
    SLBR & 26.13  & 0.908  & 15.16  & 18.07  & 0.139  \\
    LaMa & 17.97  & 0.677  & 37.33  & 44.92  & 0.326  \\
    \rowcolor{gray!15}
    MAT & 12.24 & 0.615 & 72.27 & 94.98 & 0.321\\
    DENet & 24.74  & 0.894  & 16.93  & 20.25  & 0.171  \\
    \rowcolor{gray!15}
    CoordFill & 22.66 & 0.819 & 22.69 & 37.64 & 0.149 \\
    SCATCL & 16.53 & 0.605 & 42.54 & 56.12 & 0.394 \\
    \midrule
    \rowcolor{gray!15}
    Ours \# 1 & 26.38 & 0.922 & 15.34 & 18.30 & 0.097\\
    Ours \# 2 & 25.99  & 0.920  & 15.94  & 19.02  & 0.103  \\
    \rowcolor{gray!15}
    \textbf{Ours} & \textbf{26.81}  & \textbf{0.924}  & \textbf{15.11}  & \textbf{18.01}  & \textbf{0.094}  \\
    \midrule
    \midrule
    \multicolumn{6}{l}{Performance with fixed and coarser mask per image} \\
    \midrule
    \rowcolor{gray!15}
    WDNet & 24.37  & 0.887&	18.83&	20.45& 	0.166  \\
    SplitNet & 25.53 & 0.899 & 18.11 & 19.02 & 0.143\\
    \rowcolor{gray!15}
    SLBR & 26.09 & 0.907 & 15.16 & 15.91 & 0.141\\
    LaMa & 14.87 &	0.551&	52.20	&56.50	&0.470  \\
    \rowcolor{gray!15}
    MAT & 8.31 &	0.483	&108.30&	123.88 &	0.386\\
    DENet & 25.06 & 0.900 & 16.31 & 17.14 & 0.162\\
    \rowcolor{gray!15}
    CoordFill & 17.07&	0.568&	40.52	&46.97&	0.381 \\
    SCATCL & 13.89&	0.487&	63.25&	55.00&	0.536 \\
    \rowcolor{gray!15}
    \textbf{Ours} & \textbf{26.66}  & \textbf{0.924}  & \textbf{15.09}  & \textbf{16.48}  & \textbf{0.094}  \\
    \bottomrule
    \end{tabular}%
    }
    \caption{Experimental results of different models on ILAW. \# 1: our method using unaugmented masks during training. \# 2: our method without using pretrained LaMa.}
  \label{tab2}%
\end{table}%

\noindent\textbf{$\bullet$ Real-world Dataset.}  We collect 27 high-resolution watermarked images from the Internet, and then employ LabelMe ~\cite{russell2008labelme} to draw coarse masks. 
User study is conducted for assessment on this dataset.


\subsection{Implementation Details \& Evaluation Metrics}

The codes are implemented by PyTorch \cite{paszke2019pytorch}, PyTorch-Lightning \cite{falcon2019pytorch} and Hydra \cite{yadan2019hydra}. We employ Adam optimizer \cite{kingma2014adam} with a learning rate of 0.0001 to train both generator and discriminator. The model is trained for 100 epochs with a batch size of 16. For the weights for individual sub-losses, we experiment with variation on weight factors, and observe subtle performance fluctuation. 
Finally, we set: $\omega_1 = 10, \omega_2 = 30, \omega_3 = 1, \omega_4 = 100,\omega_5 = 0.001$. For comparison, we train other models on ILAW with an additional input of coarse mask $\mathbf M$ concatenated to the original input.

To evaluate the removal efficacy, we use PSNR, SSIM, RMSE, RMSE$_w$ and LPIPS \cite{zhang2018unreasonable}. The RMSE$_w$ is RMSE averaged inside mask. LPIPS evaluates perceptually similarities in large-area content recovery.

\subsection{Comparison with Existing Methods}

\subsubsection{Visible Watermark Removal Guided by Coarse Mask}
\begin{figure}[t]
\begin{center}
    \includegraphics[width=0.6\linewidth]{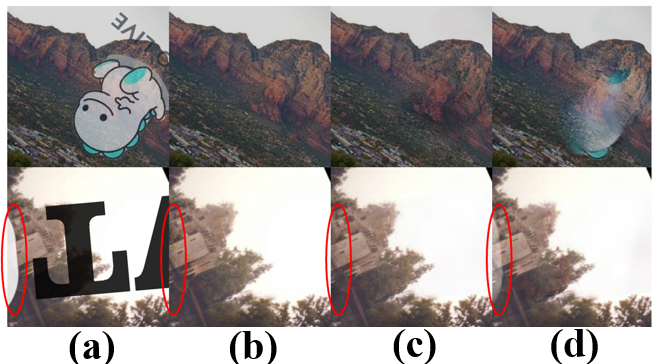}
\end{center}
   \caption{Comparisons of model without pre-training/model with pre-training. (a) Input, (b) Target, (c) Our output, (d) Output without pre-trained LaMa.
   }
\label{fig8}
\end{figure}

First, we conduct experiments on the ILAW dataset using coarse masks, evaluating methods such as WDNet \cite{liu2021wdnet}, SplitNet \cite{cun2021split}, SLBR \cite{liang2021visible}, DENet \cite{sun2023denet}, LaMa \cite{suvorov2022resolution}, MAT \cite{li2022mat}, CoordFill \cite{liu2023coordfill}, and SCATCL \cite{zuo2023generative}. We also test our approach in three scenarios: training with unaugmented masks (Ours \# 1), training without pre-training LaMa (Ours \# 2), and training with the original design (Ours). The results in the upper section of Table~\ref{tab2} show our method's superior performance, driven by the substantial prompt information extracted by WCC and BCE, which helps LaMa accurately recover background content. Ours \# 1 highlights the benefits of coarse masks for restoration, while Ours \# 2 confirms that pre-trained LaMa parameters assist in content generation. Figure~\ref{fig8} further illustrates that models without pre-training leave residual watermarks in the restored images.

\begin{figure*}[t]
\begin{center}
    \includegraphics[width=\linewidth]{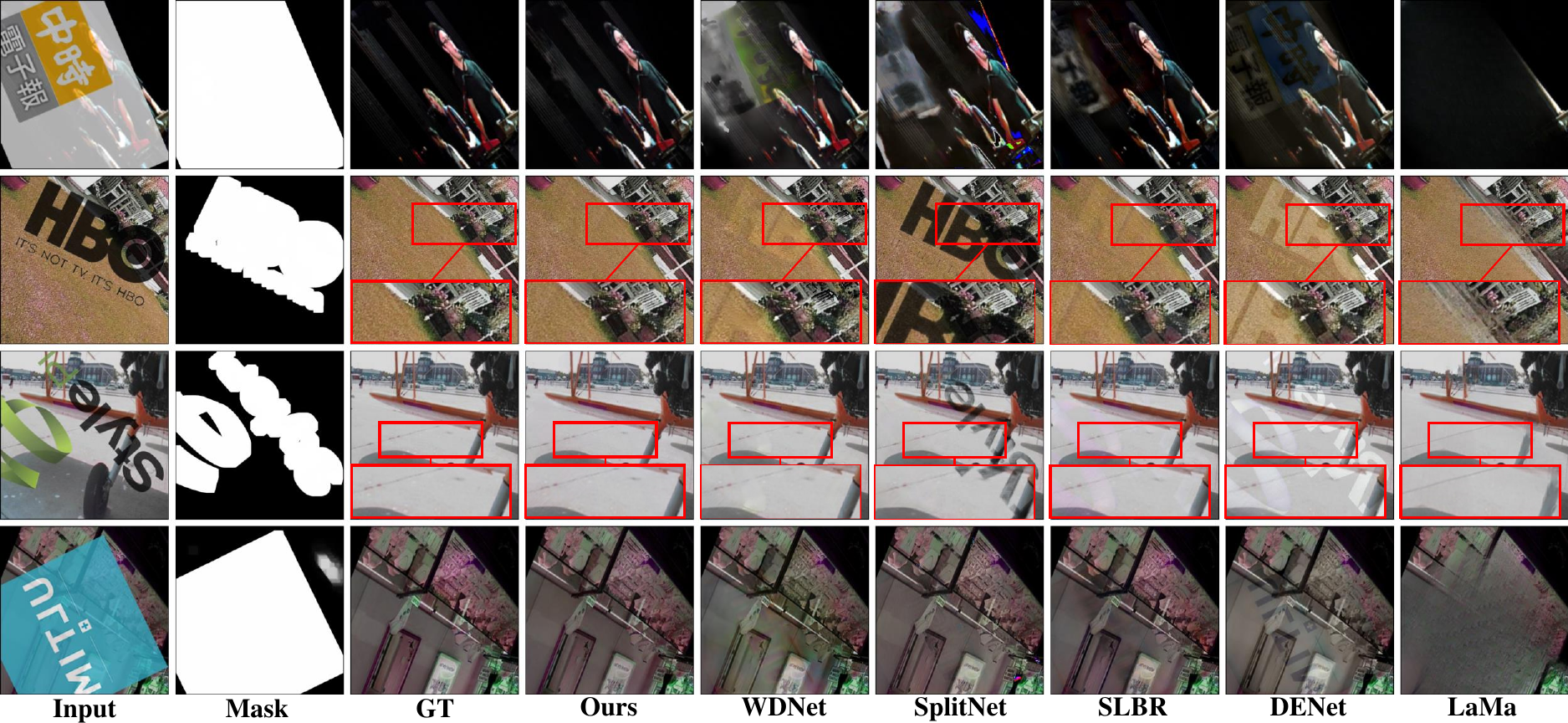}
    \end{center}
   \caption{Visualization results of different methods on ILAW.
   }
\label{fig6}
\end{figure*}

\begin{table}[h]
  \centering
  \renewcommand{\arraystretch}{0}
   \begin{tabular}{cccc}
    \toprule
    Count & SplitNet & SLBR  & \textbf{Ours} \\
    \midrule
    Vote  & 154   & 206   & \textbf{882} \\ \midrule
    Best-Restored  & 1 & 2 & \textbf{24} \\
    \bottomrule
    \end{tabular}%
    \caption{User study on the real-world dataset.}
    \label{tab6}
\end{table}%

Moreover, in the lower section of Table~\ref{tab2}, we record some testing results using coarser masks generated from augmentation with more dilation and erosion, or rough outline with polygonal shape. The reuslts demonstrates our method's robustness to low-quality masks, maintaining superior performance compared to other models.

\begin{figure}[t]
\begin{center}
    \includegraphics[width=\linewidth]{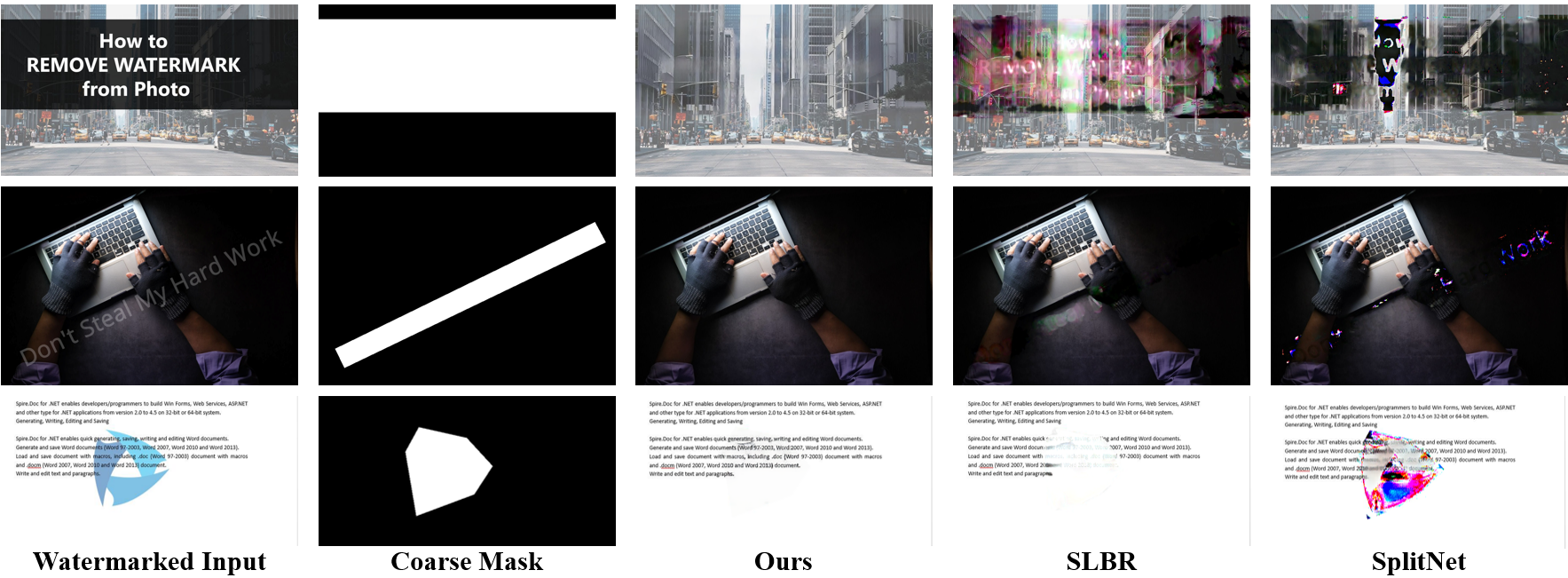}
\end{center}
   \caption{Visualization results of different methods on pictures from real-world dataset.
   }
\label{fig7}
\end{figure}
For real-world validation, we collect restoration results from the top 3 models (our model, SplitNet, SLBR) and conduct a survey with 46 participants. Each participant votes on the most effective model for each of 27 images. Table \ref{tab6} shows the total votes and the number of images where each model was preferred. The results indicate a clear consensus of our model's superiority over the alternatives.

Figures \ref{fig6} and \ref{fig7} showcase visual comparisons between our method and existing approaches. It is evident that watermark removal methods like WDNet, SplitNet, SLBR, and DENet fail to completely remove watermarks, and LaMa's inpainting produces inaccurate content within masked areas. In contrast, our model achieves superior restoration. 

\begin{table}[h]
\centering
\renewcommand{\arraystretch}{0.1}
\setlength{\tabcolsep}{1.9mm}{
    \begin{tabular}{cccccc}
    \toprule
          & WDNet & SplitNet & SLBR   & DENet & \textbf{Ours} \\
    \midrule
    PSNR$\uparrow$  & 24.37  & 25.72  & 25.02   & 19.66  & \textbf{25.77} \\
    SSIM$\uparrow$& 0.887  & 0.892  & 0.890    & 0.814  & \textbf{0.916} \\
    LPIPS$\downarrow$ & 0.166  & 0.156  & 0.154   & 0.236  & \textbf{0.100} \\
    \bottomrule
    \end{tabular}%
    }
    \caption{Performance of methods using none/white masks.}
    \label{tab:exper}
\end{table}%

\subsubsection{Blind Watermark Removal}
We also test some methods on blind watermark removal, where coarse mask inputs are absent or replaced by white masks. As shown in Table~\ref{tab:exper}, our method outperforms competitors, demonstrating its effective watermark locating even without assistance of mask input.


\subsection{Ablation Studies}

\subsubsection{Effect of Key Structures}
To demonstrate the effectiveness of our inpainting model with dual-branch feature adaptation, we conduct ablation experiments with different branch configurations. First, we train with only LaMa, as shown in the first row of Table \ref{tab3}, highlighting the need for feature adaptation, as the model alone struggles to learn features in masked regions, limiting background recovery. Next, we test a single-branch adaptation (second row of Table \ref{tab3}), which performs suboptimally due to difficulty distinguishing low-opacity watermarks from the background. In contrast, dual-branch adaptation successfully separates the watermark, allowing for better restoration. We then use two branches with transposed attention modules (third row of Table \ref{tab3}), emphasizing the importance of LaMa's long-range feature capture for large-area restoration. Finally, we test on feature adapting network with different number of blocks, and found that out 3-block structure reaches the best results.


    
\begin{table}[t]
  \centering
  \renewcommand{\arraystretch}{0.5}
  \setlength{\tabcolsep}{1.5mm}{
    \begin{tabular}{c|c|c|c}
    \toprule
    Backbone & Feature Adapting & PSNR & SSIM \\
    \midrule
    LaMa  & -     & 23.62 & 0.895 \\
    LaMa  & 3 TA blocks in BCE  & 25.67 & 0.916 \\
    WCC   & 3 TA blocks in BCE  &  25.82 & 0.917 \\
    LaMa  & 2 TA blocks in WCC+BCE  & 23.89 & 0.892 \\
    LaMa  & 6 TAblocks in WCC+BCE  & 22.82 & 0.874 \\
    \textbf{LaMa} & \textbf{3 TA blocks in WCC+BCE} & \textbf{26.81} & \textbf{0.924} \\
    \bottomrule
    \end{tabular}%
    }
    \caption{Ablation study of network structure.}
  \label{tab3}%
\end{table}%
\begin{table}[t]
  \centering
  \renewcommand{\arraystretch}{0.5}
  \setlength{\tabcolsep}{1.5mm}{
    \begin{tabular}{c|c|c|c}
    \toprule
    Feature Extraction & Fusion & PSNR & SSIM  \\
    \midrule
    Conv (kernel = 3)  & GFM   & 24.04 & 0.895 \\
    Conv (kernel = 7)  & GFM   & 20.87 & 0.855 \\
    Conv (kernel = 5, dilation = 3) & GFM   & 22.25 & 0.864\\
    Conventional Attention & GFM   & 23.50 & 0.887\\
    Transposed Attention    & Conv  & 26.02 & 0.920 \\
    \textbf{Transposed Attention} & \textbf{GFM} & \textbf{26.81} & \textbf{0.924} \\
    \bottomrule
    \end{tabular}%
    }
    \caption{Ablation study of module selection.}
  \label{tab5}%
\end{table}%
\subsubsection{Effect of Key Modules}
To validate the effectiveness of the feature extraction and fusion modules in our design, we replaced the transposed attention and GFM modules with alternatives. The results in Table \ref{tab5} show that our chosen modules are more effective, as they better capture global background and watermark features, while the GFM's gated mechanism efficiently filters useful information for restoration. The alternative modules perform suboptimally, as handling high-resolution images with large watermarks requires addressing both long-range dependencies between watermarked and unwatermarked areas and short-range dependencies within the watermarked area. Simple local or global approaches, such as convolution or dilated convolution, are insufficient. Conventional attention modules lead to excessive and meaningless calculations, resulting in minor performance and longer inference time. 

\section{Conclusion}

In conclusion, this paper introduces an innovative feature adapting framework tailored for the challenging task of large-area visible watermark removal. The proposed framework leverages specialized components, including a watermark component cleaning branch and a background content embedding branch, both equipped with transposed attention modules for enhanced feature extraction. The integration of gated fusion modules further refines the image inpainting backbone, facilitating accurate reconstruction of watermarked regions by incorporating prompt information within the extracted features. Additionally, the model exhibits adaptability to imprecise watermark masks through the incorporation of a coarse segmentation mask. Empirical evaluations conducted on two datasets demonstrate the effectiveness of our method, showcasing its state-of-the-art performance in comparison to various existing approaches.

\section{Acknowledgments}
This work was supported by the National Natural Science Foundation of China under Grant 62376206 and 62322608.

\bibliography{sec/egbib}

\end{document}